\documentclass[10pt,twocolumn,letterpaper]{article}

 \usepackage{iccv}              
\usepackage{amsmath}
\usepackage{amssymb}
\usepackage{subcaption}
\usepackage{booktabs}
\usepackage{multirow}
\usepackage{comment}
\RequirePackage[dvipsnames]{xcolor}
\usepackage[T1]{fontenc}

\usepackage{subcaption}
\usepackage{enumitem}

\definecolor{iccvblue}{rgb}{0.21,0.49,0.74}
\usepackage[pagebackref,breaklinks,colorlinks,allcolors=iccvblue]{hyperref}

\def\paperID{8201} 
\def\confName{ICCV}
\def\confYear{2025}

\title{Visual Interestingness Decoded: How GPT-4o Mirrors Human Interests}

\author{Fitim Abdullahu and Helmut Grabner\\
Zurich University of Applied Sciences, Switzerland\\
{\tt\small \{fitim.abdullahu, helmut.grabner\}@zhaw.ch}
}

\begin{document}

\maketitle

\begin{abstract}
Our daily life is highly influenced by what we consume and see. Attracting and holding one's attention -- the definition of (visual) interestingness -- is essential. The rise of Large Multimodal Models (LMMs) trained on large-scale visual and textual data has demonstrated impressive capabilities. We explore these models' potential to understand to what extent the concepts of visual interestingness are captured and examine the alignment between human assessments and GPT-4o's, a leading LMM, predictions through comparative analysis. Our studies reveal partial alignment between humans and GPT-4o. It already captures the concept as best compared to state-of-the-art methods. Hence, this allows for the effective labeling of image pairs according to their (commonly) interestingness, which are used as training data to distill the knowledge into a learning-to-rank model. The insights pave the way for a deeper understanding of human interest. Code and materials: \small\textcolor{iccvblue}{\url{https://github.com/fiabdu/Visual-Interestingness-Decoded}}
\end{abstract}

\section{Introduction}
\label{sec:intro}

Online media data continues to expand, making it increasingly challenging to deliver relevant and engaging content to users. A key aspect of this challenge is the concept of (visual) interestingness -- capturing attention and influencing behavior, which dates back to Berlyne's work in 1949~\cite{Berlyne1949InterestConcept}. On the other hand, a vast amount of online accessible media is scraped to empower the training of foundation models in a self-supervised manner. Large Multimodal Models (LMMs), especially Language-Vision Models like GPT-4o~\cite{openai2024gpt4technicalreport}, encode human-like knowledge and perform impressively across tasks. While they can reliably categorize images or answer visual questions, their ability to recognize subjective concepts remains uncertain.

\begin{figure}[t]
  \centering
  \includegraphics[width=0.72\linewidth]{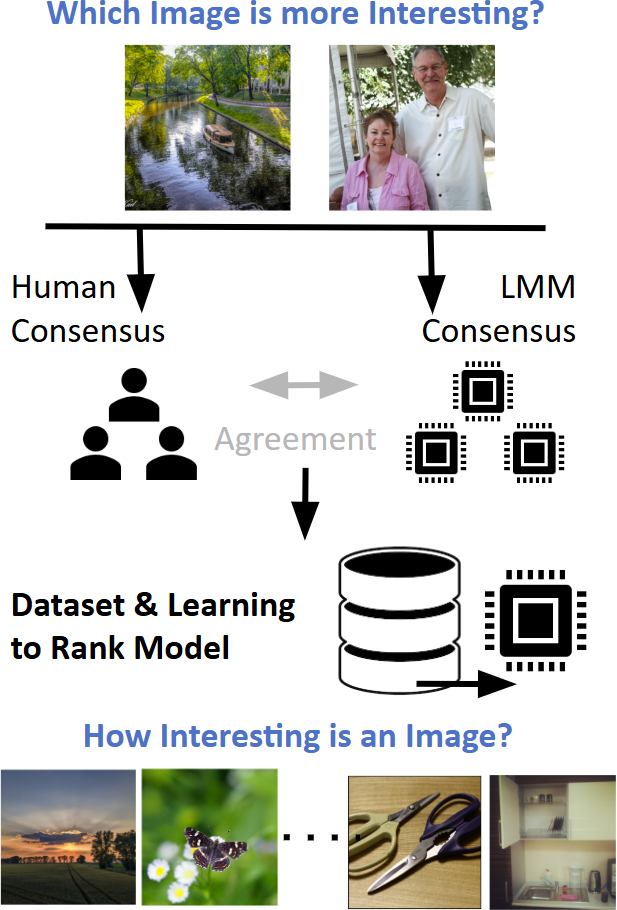}
  \caption{LMMs, such as GPT-4o, encode human-like knowledge and perform well across various tasks. We explore image interestingness, a highly subjective concept, by examining consistent labeling between humans and LMMs and their level of agreement. Pairwise labeling accesses relative measures, which are used to train a rank model to finally assess an image's interestingness.}
  \label{fig:teaser}
\end{figure}

\begin{table*}[t]
    \centering
    \small
    \begin{tabular}{p{1.25cm}||p{2.5cm}|p{4cm}|p{7.5cm}}
        \toprule
        \textbf{Approach} & \textbf{Data} & \textbf{Definition of Interestingness} & \textbf{Labels \& Computational Model} \\
        \hline
        \hline
        IJCV'21 \cite{Constantin2021VisualReview} et al. & images, image features and meta-data & explicit human annotations, e.g., AMT & ``direct'' supervised training using the human annotations \\ 
        \hline
        ECCV'24 \cite{abdullahu2024commonlyinterestingimages} & images (large scale from Flickr) & user interaction, i.e., user favorites & unsupervised estimation of user agreement to identify commonly interesting images; used to train a regressor\\ 
        \hline
        proposed & images (large scale) & LMM trained on a vast amount of (human-generated) data & label pairwise images to capture local preferences, which are then combined into a global learning-to-rank model \\ 
        \bottomrule

    \end{tabular}
    \caption{Most related work has focused on explicit human annotation. However, obtaining reliable annotations at scale is challenging and costly. Recent work explores implicit information to obtain ``real'' data from photo-sharing platforms at scale. However, this data may be biased toward a particular application, and only positive intentions (i.e., favorites) are available. To overcome these limitations, our approach leverages knowledge represented in LMMs to distill a computational model for visual interestingness.}
    \label{tab:visual_interest}
\end{table*}

This work investigates how well state-of-the-art LMMs capture the fuzzy concept of visual interestingness. We explore whether these models can identify features associated with interestingness and compare their assessments to human judgments through user studies. As illustrated in Fig.~\ref{fig:teaser}, we focus on both (i) the alignment and divergence between human- and model-based evaluations and (ii) assessing LMMs’ potential to reduce manual labeling effort by distilling knowledge from their internal representations.

We focus on everyday images to make the task traceable, consistent with prior work~\cite{deza2015virality, Dhar2011HighInterestingness, Gygli2013TheImages}. By achieving high consensus among humans, we explicitly limit subjectivity in the task, aiming to identify images that resonate with a broader audience~\cite{abdullahu2024commonlyinterestingimages, gardezi2021makes, buenzli2024}. We balanced the dataset size with experimental feasibility to ensure traceability and manage computational costs. Despite this trade-off, our results remain significant, and the insights are valid. 

\noindent To this, our main contributions are:
\setlist{nolistsep}
    \begin{itemize}[noitemsep]
  \setlength\itemsep{0em}
    \item We present a dataset of 1,000 images (2,500 image pairs) with metadata and multiple verified labels from human annotators and three state-of-the-art LMMs.
    \item We propose a novel approach to estimating image interestingness without direct user input, which outperforms state-of-the-art models for everyday images.
    \item We analyze and compare human and LMMs annotations, highlighting areas of agreement and divergence, paving the way for better understanding human and machine interests.
\end{itemize}



\section{Related Work}
\label{sec:relatedWork}

Interestingness is inherently ambiguous\footnote{Ambiguities -- i.e., missing information to specify the task explicitly -- are common~\cite{parrish-etal-2024-picture}. As a classic example, let's consider René Magritte's 1929 painting \emph{The Treachery of Images}~\cite{margritte1929pipe}. The artwork, depicting an image of a pipe with the caption "Ceci n'est pas une pipe" ("This is not a pipe"), challenges viewers' perceptions of images and symbols.}. It is subjective, varying by context, observer, and individual background~\cite{Berlyne1949InterestConcept, Gygli2013TheImages, Constantin2021VisualReview}. Yet, recent work suggests that certain image features appeal broadly, regardless of individual differences~\cite{abdullahu2024commonlyinterestingimages, buenzli2024, gardezi2021makes}.
Constantin et al.~\cite{Constantin2021VisualReview} provides a comprehensive overview of computational approaches to visual interestingness. For instance, Flickr calculates an “interestingness” score to help users discover engaging content on its platform~\cite{flickraboutint, flickrPat}. Early research in this area primarily relied on classical machine learning and computer vision methods, often constrained by limited datasets~\cite{Gygli2013TheImages, Dhar2011HighInterestingness, Grabner2013VisualSequences}. As the field has grown, the advent of deep learning and more powerful computational resources enabled larger-scale studies and more complex models, e.g.,~\cite{Constantin2019ComputationalCovariates}.

However, unbiased data collection and consistent annotations at a large scale are challenging and costly. Abdullahu and Grabner~\cite{abdullahu2024commonlyinterestingimages} recently proposed a more progressive approach that explores indirectly labeled data from multiple Flickr users. ``Interestingness'' is defined by analyzing user engagement, and their approach aims to identify commonly appealing image characteristics directly from users' favorites. In our work, we aim to extend this idea and get annotations from the knowledge encoded into LMMs trained on a vast amount of (human-generated) data -- see Tab.~\ref{tab:visual_interest}.


\textbf{Large Multimodal Models.}
The emergence of multimodal foundation models has transformed AI by combining computer vision and natural language processing within a unified framework~\cite{openai2024gpt4technicalreport, dubey2024llama3herdmodels, geminiteam2024geminifamilyhighlycapable, chen2025janus, wu2024deepseekvl2mixtureofexpertsvisionlanguagemodels}.
This paradigm shift changes training approaches and enables broader, more flexible model applications, moving from discrete classification tasks to prompt-based interactions~\cite{NIPS2013_DeViSE}. 
This flexibility is advantageous in question-answering tasks, where users may ask about unfamiliar or abstract concepts. GPT-4o achieves state-of-the-art performance on various visual perception benchmarks~\cite{openai2024gpt4technicalreport,openai2024_hello}.

Recent models like GPT-4o have been trained on vast amounts of web data, encoding extensive knowledge from different fields. Fine-tuning and distilling knowledge into smaller, more specific models is widely used; see~\cite{xu2024survey} for a recent survey. Leveraging the general knowledge of LMMs, specific models are used to design automatic evaluators that mirror human performance. Applications include evaluating text and images (e.g., 3D models)~\cite{wu2023gpteval3d}, assisting graphic design~\cite{haraguchi2024can}, assessing fashion aesthetics~\cite{YukiGPT4Fashion}, or measuring content appeal~\cite{chen2024aidappealautomaticimagedataset}.

To align with human values, models are fine-tuned with supervision~\cite{NEURIPS2022_b1efde53} or automatically~\cite{peng2024dreambench}. However, their latent knowledge can go beyond what LMMs are explicitly taught~\cite{NEURIPS2022_grokking, NEURIPS2022_locating}, potentially revealing both overlaps and gaps between human and machine understanding~\cite{iclrkeynote_been_2022, schut2023bridginghumanaiknowledgegap}. Understanding these overlaps and gaps could help uncover new concepts and insights\footnote{A widely known example might be the classical ``Move 37'' of AlphaGo when playing against Lee Sedol \cite{metz2016two}.}, especially in subjective areas.

\textbf{Outline of the Paper.} We aim to explore how the implicit knowledge encoded in LMMs relates to the concept of visual interestingness. Sec.~\ref{sec:singleImage} and Sec.~\ref{sec:twoImages} present consensus and agreement for single and relative image interestingness assessment, respectively. Sec.~\ref{sec:gloablRanking} uses automatically annotated image pairs from GPT-4o to train a computational model for predicting image interestingness within a learning-to-rank framework. Finally, Sec.~\ref{sec:assesment} discusses the similarities and differences between humans and GPT-4o, providing insights into what makes an image interesting.




\section{Single Image Interestingness Assessment}
\label{sec:singleImage}

\subsection{Experimental Design}

\textbf{\indent Dataset.} We chose images from the photo-sharing platform Flickr as these images capture a wide variety of content from diverse communities, including professionals and amateur users~\cite{OpenImages2, abdullahu2024commonlyinterestingimages}. To select representative and diverse images, we selected 1,000 images from the Flickr-User dataset~\cite{abdullahu2024commonlyinterestingimages}, equally sampled according to their proposed commonly interestingness score. 

\textbf{Human Annotations.} We conducted our user study using Amazon Mechanical Turk. Human workers were instructed on evaluating image interestingness and describing their choice, with no right or wrong answers -- it was their decision if an image matched their interest (intentionally kept very open). More specifically, a randomly selected image was shown along with the question ``Is this image interesting to you?'' with two response options: yes or no. Additionally, they were prompted to provide a brief explanation for their choice. Five Human Intelligence Tasks (HITs) were created for each of the 1,000 images. Each HIT contained one image, and workers were compensated \$0.01 per completed task. A total of 258 unique workers participated, with each worker labeling images without seeing the same image more than once.


For further analysis, we split the dataset $\mathcal{S} = \mathcal{C}_H \cup \mathcal{D}_H$. 
An image belongs to $\mathcal{C}_H$ if four or five participants ($(\cdot)_H$, for human) agree on their response, indicating a high level of consensus. Otherwise, it belongs to $\mathcal{D}_H$, reflecting dissent. Additionally, we define the interestingness label $y_H$ as 1 if the majority finds the image interesting and 0 otherwise.


\textbf{LMM Annotations.} The rapid development of foundation models is remarkable, especially considering how quickly new models are released. In our work, we use state-of-the-art LMMs, specifically OpenAI's GPT-4o~\cite{openai2024gpt4technicalreport} ('gpt-4o-2024-11-20'), Meta's Llama 3.2~\cite{dubey2024llama3herdmodels} ('Llama-3.2-11B-Vision-Instruct'), , and DeepSeeks Vision-Language VL2~\cite{wu2024deepseekvl2mixtureofexpertsvisionlanguagemodels} (`DeepSeek-VL2-tiny'). All of which can take text-image inputs and produce text outputs. The models are given the same prompt as the human annotators (the exact prompt is provided at the beginning of the subsequent subsections), with each image being evaluated five times. Analog to human annotations, consistently labeled images belong to sets $\mathcal{C}_G$ (for GPT-4o), $\mathcal{C}_L$ (for Llama 3.2), and $\mathcal{C}_D$ (for DeepSeek) others to the corresponding dissent sets $\mathcal{D}_{\{G, L, D\}}$. Annotation labels $y_{\{G, L, D\}} \in \{0, 1\}$ are defined by majority vote. 

\begin{figure}[t!]
    \centering
    \subcaptionbox{%
    \footnotesize
        \textbf{Human:} 5/5 yes; Explanations: ``ILIKE'', ``It is good photo and interesting'', ``It is amazing'', ...\\
        \textbf{GPT-4o:} 5/5 yes; Explanations: ``The vibrant sunset and scenic landscape create a captivating visual appeal.'', ``The sunset and landscape create a visually stunning scene.'', ``The sunset with radiant clouds over a vast field creates a captivating and serene scene.'', ...}
        [0.47\linewidth]{\includegraphics[width=0.95\linewidth]{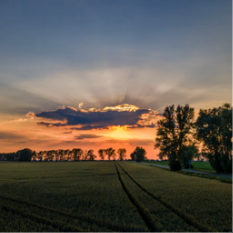}}
    \hfill
    \subcaptionbox{%
    \footnotesize
        \textbf{Human:} 5/5 yes; Explanations: ``NEAT'', ``Thinking'', ...\\
        \textbf{GPT-4o:} 4/5 yes; Explanations: ``The cluttered workshop with a person lounging hints at a story.'', ``It depicts a unique juxtaposition of a living space and a workshop.'', ``The image depicts a person relaxing in a cluttered garage, \emph{which might not appeal to everyone}.'', ...\label{fig:singleImage2}}
        [0.47\linewidth]{\includegraphics[width=0.95\linewidth]{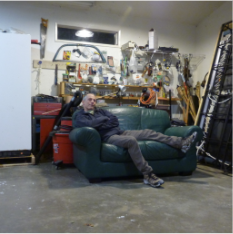}}
     
  \caption{Examples of images and corresponding responses from human annotators and GPT-4o. Almost all images are consistently labeled, most of them as interesting.}
  \label{fig:singleImage}
\end{figure}

\textbf{Asking for Explanations.} Asking ``why'' is used twofold, (i) to ensure quality for the judgments (see discussion for Llama 3.2 and DeepSeek annotations in Sec.~\ref{sec:experimental-design}) and (ii) to gain deeper insights about how humans and LMMs come to the particular conclusion 
whether the image is interesting (see Sec.~\ref{sec:assesment}). Examples are shown in Fig.~\ref{fig:singleImage}.

\subsection{``Is \underline{This} Image Interesting?''}
\noindent Prompt: \textit{Is this image interesting? Answer with one word (yes or no) without punctuation and in lowercase. Add a semicolon without space. Explain why in one sentence without going into detail.}

\begin{table}[]
    \centering
    \small
    \begin{tabular}{l|c|c|c|c}
             \toprule
         & $|\mathcal{C}_x|$ & $y_x=1$ & $A^{(H,x)}(\mathcal{S})$ & $A^{(H,x)}(\mathcal{C}_H)$ \\
        \hline\hline
        \textbf{H}uman & 91.9 \% & 99.9 \% & - & - \\
        \textbf{G}PT-4o & 93.9 \% & 95.3 \% & 92.9 \% & 93.6 \% \\   
        \textbf{L}lama & 93.1 \% & 99.8 \% & 97.1 \% & 98.3 \% \\
        \textbf{D}eepSeek & 76.2 \% & 81.4 \% & 75.3 \% & 77.3 \% \\
        \bottomrule
    \end{tabular}
    \caption{Consistency and agreement with human annotation for single image interestingness assessment. Unfortunately almost all images are consistently labeled as interesting.}
    \label{tab:singleImage}
\end{table} 
As summarized in Tab.~\ref{tab:singleImage}, the annotations from humans and LMMs show high consistency (almost all images are in the respective consistent set $\mathcal{C}_x$). This indicates that humans and LMMs generally agree on the interestingness of images.
Furthermore, almost all images in the respective sets are considered interesting ($y_x = 1$), indicating that humans and LMMs find almost all images on which they agree to be interesting.
These suggest that humans and LMMs actively look for something interesting when explicitly asked for it, leading to a predominantly positive response.

The agreement $A^{(M,N)}(\mathcal{S}):= \frac{1}{|\mathcal{S}|} \sum_{i=1}^{|\mathcal{S}|} \mathbb{I}(y^{(i)}_{M} = y^{(i)}_{N})$ between annotation $M$ and $N$ on set $\mathcal{S}$ measures how well annotations of humans and LMMs are aligned. 
Not surprisingly, as almost all images are interesting, the results indicate a high level of agreement between humans and LMMs.
Furthermore, the agreement between human and LMM increases slightly when focusing on consistently labeled images in $\mathcal{C}_H$.
It seems that it is somewhat easier for the LMM to distinguish between interesting and uninteresting if the humans agree on this question.

Notably, the DeepSeek model has significantly less consistency (among itself) and less agreement with humans. However, for humans and the other LMMs, some images fall into the dissenting sets. This inconsistency may arise because an image may not interest a broad audience. For example, in Fig.~\ref{fig:singleImage2}, GPT-4o stated: ``The image depicts a person relaxing in a crowded garage, \emph{which might not appeal to everyone}.'' This response suggests that the model subjectively evaluates the content to determine its interestingness, as discussed in~\cite{abdullahu2024commonlyinterestingimages}.


\textbf{Key Insight.} Responses from humans, GPT-4o, and Llama 3.2 are very consistent and aligned. Almost all images were deemed interesting, and a story was made up to support the decision.

\begin{figure}[t!]

        \begin{center}
            \includegraphics[width=0.95\linewidth]{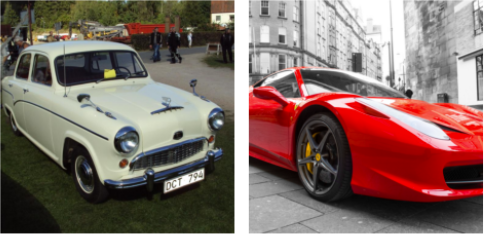}
        \end{center}

        \footnotesize
        \textbf{Human:} 5/5 second; Explanations: ``Love Ferraris!'', ``Very nice'', ``It looks good'', ...\\
        \textbf{GPT-4o:} 5/5 second; Explanations: ``The vibrant color and modern design stand out more.'', ``The modern, sleek design of the vehicle coupled with the vibrant color captures attention more effectively.'', ``It's visually striking due to its modern design and vivid color.'', ...

        \begin{center}
            \includegraphics[width=0.95\linewidth]{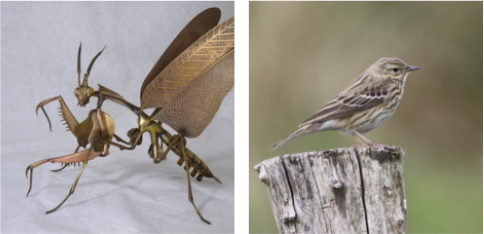}
        \end{center}
        \footnotesize
        \textbf{Human:} 5/5 second; Explanations: ``I like birds'', ``This bird is cute'', ``LOOKING NATURAL'', ...\\
        \textbf{GPT-4o:} 5/5 first; Explanations: ``The intricate design and craftsmanship make it more visually engaging.'', ``It showcases a unique and artistic representation.'', ``The first image depicts a unique and intricate metallic insect sculpture, making it more visually striking.'', ...

    \caption{Image pairs illustrating instances where humans and GPT-4o agree and disagree. For example, humans and GPT-4o have differing opinions regarding images of insects and birds. At first glance, it may not be immediately evident that the insect image is a metallic sculpture, which could explain why people did not find it as interesting—humans may not give the image the same level of attention as a machine.}
    \label{fig:img_pair_agreement}
\end{figure}

\section{Relative Image Interestingness Assessment}
\label{sec:twoImages}

As demonstrated in the last sections, whether an image is interesting is hard to answer generally. Results on single images are rendered meaningless, as almost all images are consistently labeled as interesting by humans and state-of-the-art LMMs. Relative comparisons are more affordable and often used for similar judgments, e.g., \cite{wu2024openendedvisualqualitycomparison, Gygli2013TheImages}. 

\subsection{Experimental Design}
\label{sec:experimental-design}

\textbf{\indent Dataset.} We created image pairs based on the 1,000 images used previously. Each image was used in five different (random) pairs, resulting in 2,500 image pairs. 

\textbf{Human Annotations.} As in the previous experiment, Amazon Mechanical Turk was used to obtain human annotations. A randomly selected image pair was shown to a worker, who was asked: ``Which image is more interesting to you?'' The worker could click on their preferred image and was asked to briefly explain their choice. Consistent with the previous experiment, five HITs were created for each of the 2,500 image pairs. Each HIT contained one image pair, and workers were compensated \$0.01 per completed task. A total of 553 unique workers participated, with each worker labeling image pairs without seeing the same pair more than once. 

As above, the consistency of the answers is defined if four or more humans agree on the labeling ($\mathcal{C}_H$). 
Furthermore, let $y_H$ represent the human labels, where $y_H = 1$ if the majority prefers the first image and $y_H = 0$ if the majority prefers the second image. 


\textbf{GPT-4o Annotations.} In this study, two image inputs are used, which is supported by GPT-4o.

\textit{Systematic Error.} Even though the model allows multiple images as input, we have discovered a systematic error. Image pairs were presented twice to GPT-4o, and the images were swapped on the second run. For 36\% of the cases, GPT-4o always reported the second image as more interesting. Only 64\% (1,599 out of 2,500) of the GPT-4o annotation remained the same, independent of the image order.
For subsequent experiments, only these image pairs were kept.
Please note that this systematic error does not seem to be much correlated to the human consensus (56.3\% of image pairs in $\mathcal{C}_H$ and 47.6\% in $\mathcal{D}_H$ are error-free).


\textit{GPT-4o Demographics.} As interestingness is subjective, it would be nice to test different user groups based on their demographics automatically. We used the system prompt of GPT-4o for that purpose: \textit{``You are a [gender] from [continent] and between [age] and [age] years old.''} If one uses prominent images that have become naturalized in society for men or women, such as a blue car or a pink flower, GPT-4o responds differently. A car is more interesting for men and a pink flower for women, somehow capturing prejudices. However, this vanishes when using everyday images where this distinction is no longer so prominent. In a more extensive study involving 500 random image pairs, we used \emph{male} or \emph{female} for gender and \emph{North America} or \emph{Africa} for the continent, and a range of \emph{25} to \emph{34} and \emph{45} to \emph{54} years for age, respectively. Running all eight combinations, filtering out pairs with a systematic error, and combining the remaining pairs, we ended up with 116 image pairs. Unfortunately, the results were identical for all image pairs, regardless of gender, continent, or age specified. 

\textbf{Llama 3.2 and DeepSeeks-VL2 Annotations.} As Llama 3.2~\cite{dubey2024llama3herdmodels} currently does not support multiple image inputs, we combined the two images into a single input for the model. However, this workaround did not yield reliable results. For instance, the model often selected the first image as more interesting while providing an explanation that referred to the second image. Similarly, DeepSeek's recent Janus Pro Model~\cite{chen2025janus} does not support multiple image inputs. When combining images, the model consistently selected the second image as more interesting. While DeepSeek's Vision-Language V2 model do allow for multiple image inputs~\cite{wu2024deepseekvl2mixtureofexpertsvisionlanguagemodels}, the selection and descriptions often do not align with the actual content, exhibiting issues similar to those observed with Llama 3.2.

Due to these inconsistencies, which result in an unfair comparison between these LMMs and GPT-4o, we limited our further analyses to GPT-4o. 

\textbf{Key Insight:} GPT-4o responds independently of the demographic tested; however, it has a significant systematic bias in favoring the second image over the first.

\subsection{``Which image is \underline{More} interesting?''}
\label{subsec:labelPairs}
Prompt: \textit{Which of the two images is more interesting? Answer with one word (first or second) without punctuation and in lowercase. Add a semicolon without space. Explain in one sentence why you have chosen this image without going into detail.}

Unlike in the single-image study above, people's responses are less consistent in the paired-image experiment. Set $\mathcal{C}_H$ contains 56.3\% of the image pairs, indicating consensus in about half of them. GPT-4o exhibits much higher consistency, with 95.5\% of all image pairs in $\mathcal{C}_G$. The overall agreement between GPT-4o and human annotations is $A^{(H,G)}(\mathcal{S}) = 66.2\%$. In case of human consensus, the agreement increases to $A^{(H,G)}(\mathcal{C}_H) = 73.8\%$, while decreases on the dissent set to $A^{(H,G)}(\mathcal{D}_H) = 56.5\%$ -- as one would expect. Examples are depicted in Fig.~\ref{fig:img_pair_agreement}.

\textbf{Key Insight.} 
GPT-4o’s annotations are aligned with human judgments, especially when there is consensus among people.


\begin{table*}[t]
  \centering
  \small

  \begin{tabular}{@{}l|l||cc||cc|cc@{}}
    \toprule
    Group & Model &  \multicolumn{2}{c||}{\textbf{Annotations (Sec.~\ref{sec:twoImages})}} & \multicolumn{4}{c}{\textbf{Learning to Rank (Sec.~\ref{sec:gloablRanking})}} \\
    \cmidrule(r){3-8} 
    & & $A^{(H,x)}$ & $A^{(G,x)}$ & $Acc.^{(H)}$ & $r^{(H)}_{S}$ & $Acc.^{(G)}$ & $r^{(G)}_{S}$ \\
    \midrule\midrule
    \multirow{1}{*}{\textbf{Human}} 
    & Human  & - & 73.8 \% & 77.5 $\pm$ 2.5 \%& - & 72.0 $\pm$ 3.4 \%& 0.59 $\pm$ 0.06\\
    \cmidrule(lr){1-8}
    \multirow{2}{*}{\textbf{LMMs}} 
    & GPT-4o & \textbf{73.8} \% & - & \textbf{73.4 $\pm$ 3.4 \%} & \textbf{0.59 $\pm$ 0.06} & 84.8 $\pm$ 2.5 \%& - \\
    & CuPL   & 60.3 \% & 60.9 \%  & 61.5 $\pm$ 3.5 \%& 0.34 $\pm$ 0.07 & 63.2 $\pm$ 3.1 \% & 0.42 $\pm$ 0.08\\
    \cmidrule(lr){1-8}
    \multirow{3}{*}{\parbox{2cm}{\textbf{Computational Models}}}
    & CI     & 69.6 \%  &  67.6 \% & 69.6 $\pm$ 3.6 \%& 0.54 $\pm$ 0.06 & 69.1 $\pm$ 3.3 \% & 0.52 $\pm$ 0.06\\
    & Memorability &  35.5 \%  & 39.1 \% & 34.7 $\pm$ 4.0 \% & -0.42 $\pm$ 0.08 & 38.3 $\pm$ 3.6 \%& -0.34 $\pm$ 0.07\\
    & Aesthetic   & 68.3 \% & \textbf{75.1} \% & 69.0 $\pm$ 3.7 \% & 0.50 $\pm$ 0.07 & \textbf{73.6 $\pm$ 3.7} \%& \textbf{0.67 $\pm$ 0.06} \\
    \cmidrule(lr){1-8}
    \multirow{3}{*}{\parbox{2cm}{\textbf{Social Interestingness}}} 
    & \#Views       & 61.7 \% &  63.9 \% & 63.4 $\pm$ 3.4 \%& 0.39 $\pm$ 0.08 & 66.3 $\pm$ 3.4 \%& 0.48 $\pm$ 0.08\\
    & \#Favorites   & 66.4 \% &  74.0 \% & 66.3 $\pm$ 3.2 \%& 0.47 $\pm$ 0.07 & 69.4 $\pm$ 3.1  \%& 0.57 $\pm$ 0.07\\
    & \#Comments    & 68.0 \% &  74.8 \% & 66.6 $\pm$ 3.1 \%&  0.46 $\pm$ 0.07 & 70.2 $\pm$ 3.2 \%& 0.58 $\pm$ 0.07\\
    \bottomrule
  \end{tabular}%

  \caption{
  GPT-4o achieves the highest agreement $A^{(\cdot,x)}$ among all models using human responses as the ground truth. It also outperforms existing models focused on visual interestingness, related concepts, and social metrics. On the right, we show the model's accuracy $Acc.^{(\cdot)}$ on the image pairs. The global ranking (measured by the Spearman rank correlation $r_S^{(\cdot)}$) of the test dataset remains consistent, indicating that the learning-to-rank model generalizes beyond pairwise relationships for both human and GPT-4o annotations.}
  \label{tab:agreement_metrics}
\end{table*}

\subsection{Comparisons and Relation to other Approaches}
\label{subsec:otherAnnotations}


Our annotations are compared to other models and concepts related to visual interestingness prediction  (c.f.~\cite{Constantin2021VisualReview}). All approaches provide a measurement, score, or probability per image, which (is claimed) to be related to the image's interestingness. An image pair is labeled according to which image yields the higher response.

\textbf{Aesthetics~\cite{ke2023vila}.} The VILA (Vision-Language Aesthetics) model learns image aesthetics by analyzing user comments alongside images. It models subjective aesthetic judgments by aligning visual features with language and categorizing images according to these learned aesthetics.

\textbf{Memorability~\cite{Isola2011WhatMemorable}.} A predictive model assigns memorability scores to images. Through large-scale experiments with memory recall tasks, quantified memorability is established as a stable metric across viewers.  We use the recent re-implementation from ~\cite{needell2022embracing}.

\textbf{Commonly Interesting Images (CI)~\cite{abdullahu2024commonlyinterestingimages}.} Interestingness is subjective. However, some images appeal to a broader audience and are, therefore, of common interest. A predictive model was trained by analyzing how many unique Flickr users ``favored'' images from a certain category (i.e., visually similar images).

\textbf{Social Interestingness~\cite{deza2015virality}.} Whereas being related to visual interestingness, factors beyond image features are relevant to make an image go viral. Social interestingness metrics use the number of views, favorites, and comments of a post on a social media platform. Using Flickr images in our study, we directly sourced these values for every image.   

\textbf{Zero shot learning~\cite{pratt2023does}.}
We use Customized Prompts via Language (CuPL) to generate prompts to determine an image's interestingness. Using a Large Language Model, in our case GPT-4o, to ``Describe what an \emph{interesting} image looks like'', we get various prompts such as ``An interesting image features vibrant colors, unexpected elements, and a captivating composition that draws the viewer's eye''. Overall, 500 prompts are created following~\cite{pratt2023does}. After removing duplicates and highly similar prompts, we ended up with 250
unique prompts. Text embeddings for these prompts using CLIP~\cite{radford2021learning} are calculated and averaged. The final score is the cosine similarity between text and image CLIP embeddings.

\textbf{Results and Discussion.} Results can be seen in Tab.~\ref{tab:agreement_metrics} (left). Every approach is compared to human ($y_H$) and GPT-4o annotations ($y_G$), in terms of agreement on the set $\mathcal{C}_H$. 
GPT-4o is superior to previous models in this context, followed by models using aesthetic or common interestingness. Social interestingness scores reveal that images with more comments tend to be considered more interesting than those without, which aligns with research in that regard~\cite{doi:10.1016/j.intmar.2012.01.003}. Memorability score has the weakest link to interestingness, consistent with prior findings~\cite{Gygli2013TheImages}. When GPT-4o is used as the ground truth, the VILA aesthetic model performs the best, followed by the GPT-4o model, while the human model ranks third. It also appears to have a stronger agreement with social interestingness, possibly due to their pre-training.

\textbf{Key Insight.} 
GPT-4o’s annotations are superior to previously proposed approaches to predict human interest.


\section{Learning a Computational Model}
\label{sec:gloablRanking}
\noindent So far, image pairs have been annotated by humans, GPT-4o, and various computational models. In this section, we distill this knowledge into a simple computational model.

\textbf{Learning-To-Rank.} A simple learning-to-rank model can be implemented using a Siamese network architecture with shared weights~\cite{burges2005learning}. As we are using images $\mathbf{I_0}$ and $\mathbf{I_1}$ as input, they are first embedded using CLIP\footnote{We also perform experiments with DINOv2~\cite{oquab2023dinov2} embeddings. Similar, slightly worse (67.1\% for Humans and 68.3\% for GPT-4o) results and trends were achieved. This might be because CLIP was trained on text-image pairs, which provided some supervision, whereas DinoV2 is trained purely in a self-supervised manner on images. For more details, see the supplementary material.}, passing through a linear layer with a single neuron. The scoring function is the difference between them, passed through a sigmoid function:, i.e., $S(\mathbf{I_0}, \mathbf{I_1}) := \sigma (\mathbf{w}^\intercal \mathrm{CLIP} (\mathbf{I_0}) - \mathbf{w}^\intercal \mathrm{CLIP} (\mathbf{I_1}) )$. Learning is done to maximize the score differences between pairs. Binary cross-entropy loss on the target $y \in \{0, 1\}$ is used; $y=1$ if $\mathbf{I_0}$ ranks higher than $\mathbf{I_1}$ and $y=0$ otherwise. As the weights are shared, after training, a score can also be obtained using a single input $S(\mathbf{I}) = \sigma (\mathbf{w}^\intercal \mathrm{CLIP} (\mathbf{I}))$. For multiple images, the individual scores are used to rank them. Besides its simplicity, this approach has been used successfully many times, also for distilling information from LLM or LMM, e.g., \cite{wu2024openendedvisualqualitycomparison, chen2024aidappealautomaticimagedataset}

\textbf{Training/ Testing.} The dataset was split in half for training and testing. We train learning-to-rank models for all annotations (human and GPT-4o and approaches from Sec.~\ref{subsec:otherAnnotations}). Each model was trained for 25 epochs, and no overfitting was observed. Each experiment was repeated 50 times with different training/ test splits. The results are depicted in Tab.~\ref{tab:agreement_metrics} (right). $Acc.$ denotes the model accuracy on the individual image pairs (as trained) and $r_S$ the Spearman rank correlation on the global ranking based on the scores $S(\cdot)$ for each image. 

\textbf{Results and Discussion.}
The best-performing model is obtained when training and test data are from the same source (human or GPT-4o). This serves as the baseline for comparing the other models. All the results match nicely with those from the previous sections (left side of the table) for the individual performance of labeling image pairs. 

According to human annotations, the model generalizes to unseen data, although the average accuracy of approximately 77.5\% may reflect the subjectivity of the task. GPT-4o achieves the best performance among the models, although a gap remains compared to the baseline. Other computational models, such as CI or aesthetics, perform well but still fall short of GPT-4o's results. Examining the Spearman correlations, we find that human responses positively correlate with all models except memorability, which aligns with current research findings. Notably, the correlation between humans and GPT-4o is 0.59, indicating a moderate positive relationship between these two variables.

Based on GPT-4o annotations, the aesthetic model ranks highest in accuracy after GPT-4o itself, followed closely by the human model. The global ranking of the test dataset is inline, meaning that the learning-to-rank model can generalize beyond pairwise relationships for both human and GPT-4o annotations. 

\textbf{Int10k Dataset.} 
The Int10k dataset~\cite{Constantin2021VisualReview} focuses on video summarization, and there is a significant domain gap between {\it single everyday images} (our focus in this work) and cinematic image sequences. Nevertheless, we applied our approach. Overall, the results show a significant drop, with accuracy for the human model and human-provided annotations being around 59.3\% $\pm$ 2.5\%. Even for this dataset, GPT-4o outperformed all other approaches, achieving a comparable accuracy of 59.2\% $\pm$ 2.1\%. For more details, see the supplementary material.

\textbf{Key Insight.} Computational models obtained from distilling the information from the annotations match them very well. GPT-4o is superior to previous models.


\begin{figure}
    \centering
    \begin{subfigure}[a]{\linewidth}
        \begin{center}
        \includegraphics[width=\columnwidth]{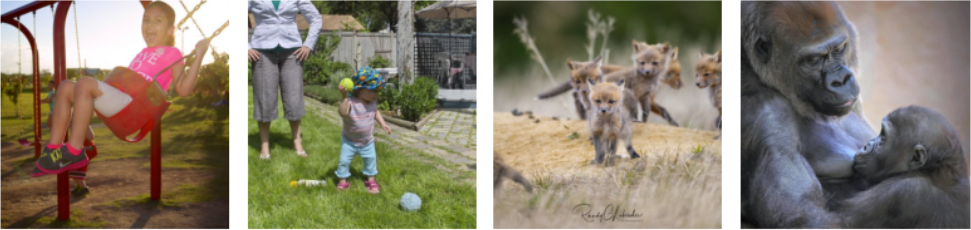}
        \end{center}
        \footnotesize
        \textbf{Human:} ``Cute Girl.'', ``This side is cute'', ``THIS IMAGE IS VERY BEAUTIFULL'', ``It looks interesting''\\
        \textbf{GPT-4o:} ``It captures a joyful moment.'', ``It shows human interaction and activity.'', ``The image of the baby animals
  captures a unique and lively moment.'', ``The emotional connection between the two beings adds an engaging element.''
  
    \begin{center}
        \includegraphics[width=\columnwidth]{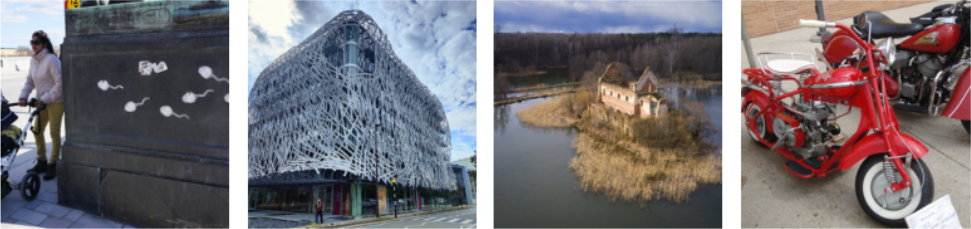}
        \end{center}
        \footnotesize
\textbf{Human:} ``Right is better than left'', ''This building's design is very interesting.'', ``Appreciate nature more'', ``What a two wheeler excited to ride immediately''\\
        \textbf{GPT-4o:} ``Features unique graffiti on a monument.'', ``The building's unique architecture makes it stand out.'', ``The image depicts a unique and intriguing structure situated in a   picturesque and seemingly remote location.'', ``The vintage motorcycle has a unique and nostalgic appeal.''
  
        \caption{Clusters considered as ``interesting''  for humans and GPT-4o.}
        \label{fig:why_positive}
    \end{subfigure}
    \vspace{1em}
    
    \begin{subfigure}[a]{\linewidth}
        \begin{center}
        \includegraphics[width=\columnwidth]{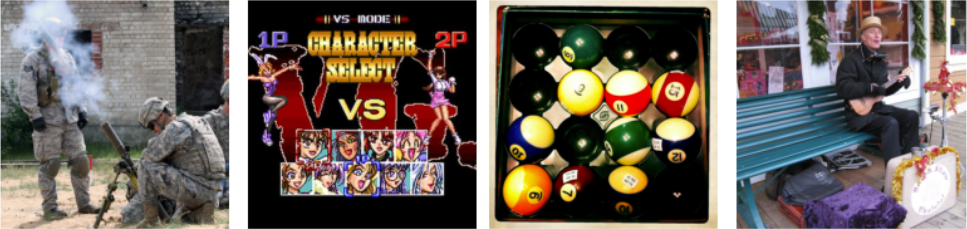}
        \end{center}
        \footnotesize
        \textbf{GPT-4o:} ``The action and dynamic scene with soldiers makes it more interesting.'', ``It has more vibrant colors and dynamic elements'', ``The arrangement and colors are visually appealing.'', ``The image shows a person performing music on the street which adds more dynamic and action.''

        \caption{Only ``interesting'' for GPT-4o (no human descriptions are available, as these images were never considered interesting in pairwise comparisons by human annotators).}
        \label{fig:why_negative}
    \end{subfigure}
    
    \caption{Explanations: The clusters are derived from text embeddings of responses "Why" an image is interesting. Clusters of common interest (a) include ``cute'', ``joyful'' moments as well as  ``uniqueness''. A minority of clusters (b) shows disagreement between humans and GPT-4o.}
    \label{fig:reasoning_examples}
\end{figure}

\section{Similarities and Differences in Assessment}
\label{sec:assesment}
This section discusses the agreement and disagreement between humans and GPT-4o in more detail.

\textbf{Embeddings.} We analyzed the responses from both GPT-4o and human annotators to create a semantic embedding using OpenAI's `text-embedding-3-small' model (1536-dimensional vector for each text input). Since the human responses were predominantly short and repetitive -- often consisting of simple terms like ``nice'', ``beautiful'', or ``good'' -- they were less suitable for in-depth analysis (cf. Fig. \ref{fig:reasoning_examples}). Therefore, we focused our analysis on the text embeddings from GPT-4o responses.

\textbf{Exploiting the ``Why'' responses.}
We perform hierarchical clustering on the embeddings. Several clusters emerged from the analysis, including the \textit{cute} and \textit{emotional} clusters shown in Fig. \ref{fig:why_positive}, which are of interest to both humans and GPT-4o. While humans respond to certain images as \textit{cute}, GPT-4o tends to associate them with emotions. Additionally, humans and GPT-4o demonstrate consensus regarding \textit{uniqueness}. Nevertheless, there are clusters that only GPT-4o finds interesting, such as images featuring \textit{vibrant color} or depicting action. These images are not necessarily captivating for humans, see Fig.~\ref{fig:why_negative}. Please note that these clusters represent semantically similar texts, not semantically similar images.


\begin{figure*}[t!]
    \centering
    \begin{subfigure}{0.76\linewidth}
        \centering
        \includegraphics[width=\linewidth]{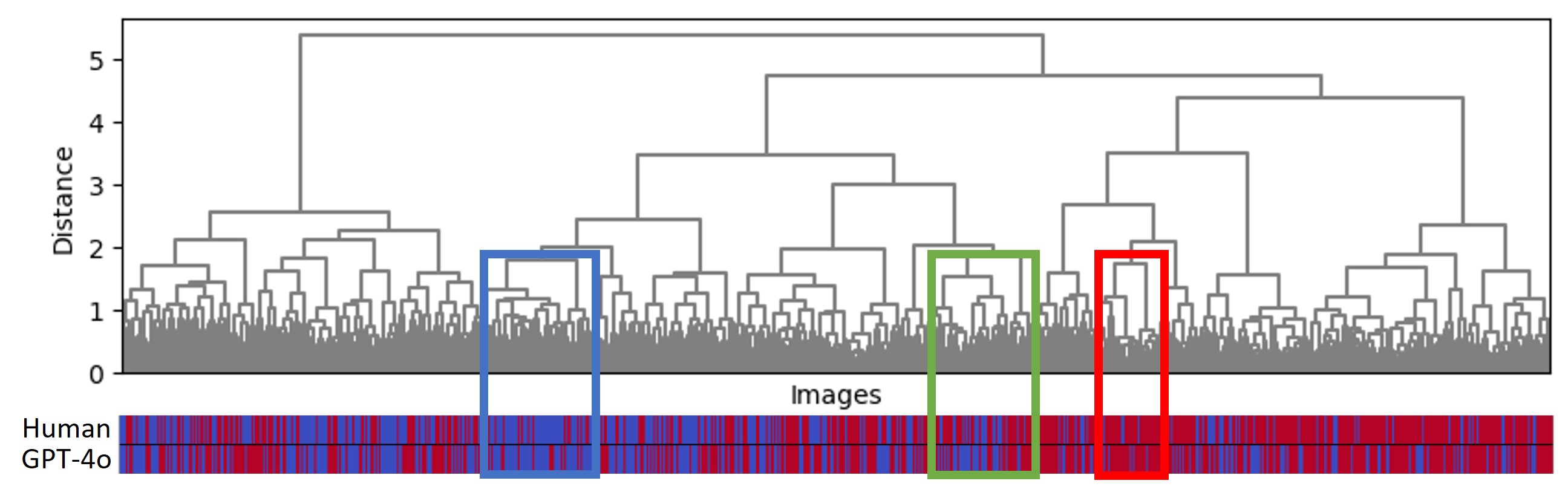}
        \caption{Hierarchical clustering reveals groups of semantically similar descriptions $(d < 2)$, each containing either exclusively interesting (\textcolor{red}{red}), uninteresting (\textcolor{blue}{blue}), or a mix of differing images (\textcolor{ForestGreen}{green}).}
        \label{fig:text_embeddings_hierarchical}
    \end{subfigure}
    \hfill
    \begin{subfigure}{0.22\linewidth}
        \centering
        \includegraphics[width=\linewidth]{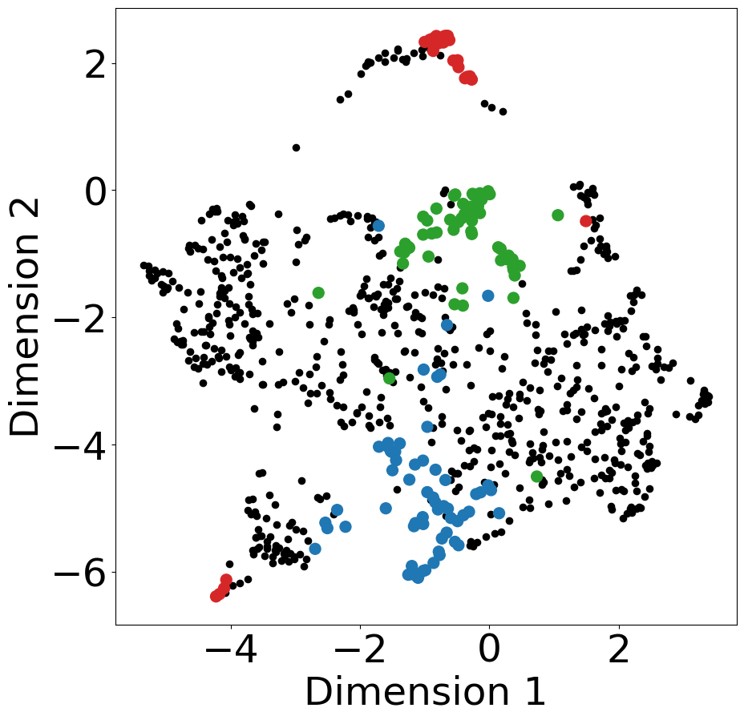}
        \caption{2d visualization of the description text, embeddings using UMAP.}
        \label{fig:text_embeddings_umap}
    \end{subfigure}
    \vskip\baselineskip
    \begin{subfigure}{0.32\linewidth}
        \centering
        \includegraphics[width=\linewidth]{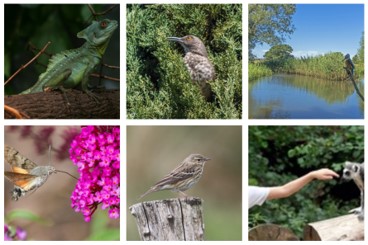}
        \caption{Agreement on interestingness (\textcolor{red}{red})}
        \label{fig:hierarchical_int}
    \end{subfigure}
    \hfill
    \begin{subfigure}{0.32\linewidth}
        \centering
        \includegraphics[width=\linewidth]{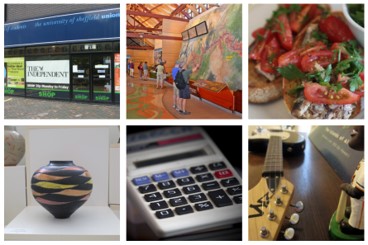}
        \caption{Agreement on uninterestingness (\textcolor{blue}{blue})}
        \label{fig:hierarchical_n_int}
    \end{subfigure}   
    \hfill
    \begin{subfigure}{0.32\linewidth}
        \centering
        \includegraphics[width=\linewidth]{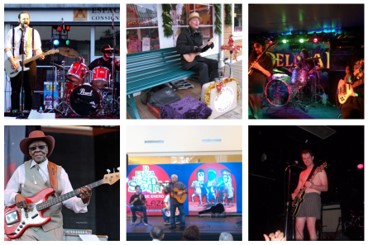}
        \caption{Disagreement (\textcolor{ForestGreen}{green})}
        \label{fig:hierarchical_mixed}
    \end{subfigure}
    \caption{Image Content Descriptions: Text embeddings of the image descriptions are divided into semantically similar groups using hierarchical clustering (a, b). Most clusters indicate agreement between humans and GPT-4 (c, d), while some indicate disagreement (e). Compare with Tab.~\ref{tab:cluster_agreement_sorted}.}
    \label{fig:GT_image_samples_combined}
\end{figure*}

\textbf{Exploiting Image Appearance.}
To gain deeper insight into what makes an image interesting or uninteresting, we conducted an additional experiment focusing on the visual characteristics of the images. GPT-4o was asked to describe each image, and embeddings were calculated for those descriptions. 
Fig.~\ref{fig:text_embeddings_hierarchical}-~\ref{fig:text_embeddings_umap} illustrates a hierarchical clustering of the description embeddings in the original space and a 2d projection using UMAP~\cite{mcinnes2018umap}. The clustering reveals groups of semantically similar images consistently found interesting, uninteresting, or divisive by humans and GPT-4o, see Fig.~\ref{fig:hierarchical_int}-\ref{fig:hierarchical_mixed}. There is agreement on images depicting flowers, birds, or nature scenes as interesting. In contrast, humans and GPT-4o find images of mundane scenes or people in ordinary situations uninteresting. Images depicting people at events or performing art elicit mixed responses.

Tab.~\ref{tab:cluster_agreement_sorted} reports quantitative information per cluster, including agreement between humans and GPT-4o and if the cluster is mainly interesting or uninteresting. Additionally, we calculated the mean ranks of images in each cluster based on both human ($\overline{R}^{(H)}$) and GPT-4o ($\overline{R}^{(G)}$) annotations. The mean ranks align closely, particularly when agreement is high. The top four words describe the cluster obtained after removing stop words and lemmatizing the automatically generated appearance descriptions. These words align well with examples from Fig.~\ref{fig:GT_image_samples_combined} (marked). 

\textbf{Key Insight.} Humans and GPT-4o generally agree well on what is interesting for topics and scenes.





\begin{table}
  \centering
  \small
  
  \begin{tabular}{@{}c|cccl@{}}
    \toprule
    $A$ & $\frac{\#A_{pos}}{\#A}$ & $\overline{R}^{(H)}$ & $\overline{R}^{(G)}$ & Frequent Words (Appearance) \\
    \midrule
    \midrule
     94\% & 60\% & 143 & 184 & {train, track, station, railway} \\
     86\% & 23\% & 246 & 236 & {people, group, front, standing} \\
     84\% & 86\% & 86 & 107 & {water, sky, background, body} \\
     \textcolor{red}{\textbf{82\%}} & \textcolor{red}{\textbf{97\%}} & \textcolor{red}{\textbf{53}} & \textcolor{red}{\textbf{80}} & \textcolor{red}{\textbf{{perched, bird, branch, flower}}} \\
     81\% & 24\% & 217 & 207 & {person, sitting, room, window} \\
     81\% & 59\% & 165 & 142 & {building, modern, street, large} \\
     80\% & 46\% & 204 & 193 & {people, two, smiling, together} \\
     80\% & 78\% & 125 & 131 & {dog, lying, person, cat} \\
     78\% & 88\% & 72 & 79 & {tree, sky, landscape, water} \\
     \textcolor{blue}{\textbf{77\%}} &  \textcolor{blue}{\textbf{18\%}} &  \textcolor{blue}{\textbf{223}} &  \textcolor{blue}{\textbf{221}} &  \textcolor{blue}{\textbf{{various, small, featuring, ...}}} \\
     77\% & 41\% & 202 & 191 & {person, wearing, standing, red} \\
     72\% & 44\% & 212 & 158 & {person, people, background, ...} \\
    72\% & 89\% & 74 & 91 & {water, bird, swimming, white} \\
    71\% & 61\% & 197 & 158 & {people, group, person, flag} \\
    74\% & 54\% & 154 & 185 & {car, parked, red, background} \\
    65\% & 72\% & 95 & 160 & {flower, green, yellow, pink} \\
    47\% & 71\% & 207 & 163 & {playing, stage, performing, guitar} \\
    \color{ForestGreen}{ \textbf{40\%}} & \textcolor{ForestGreen}{\textbf{50\%}} & \textcolor{ForestGreen}{\textbf{207}} & \textcolor{ForestGreen}{\textbf{163}} & \textcolor{ForestGreen}{\textbf{{stage, guitar, person, group}}} \\
     \bottomrule
  \end{tabular}%

  \caption{Agreement of humans and GPT-4o concerning the image content; agreement ($A$) on \textcolor{blue}{un}- and \textcolor{red}{interestingness}, average ranks form models trained using  human and GPT-4o annotations as well as \textcolor{ForestGreen}{disagreement} along its clusters text descriptions. Color markings match Fig.~\ref{fig:GT_image_samples_combined}.
 }
  \label{tab:cluster_agreement_sorted}
\end{table}

\section{Conclusion}
\label{sec:concusion}
GPT-4o cannot be used directly to assess image interestingness due to uninformative (almost always positive) responses on single images and systematic errors. However, we demonstrated that GPT-4o outperforms existing models in a comparative annotation setting. Its alignment with human assessments highlights its potential to support (i) large-scale studies and (ii) knowledge distillation. Further investigation into the gap between GPT-4o and human annotation, including demographic factors, would be insightful. Extending the study to datasets beyond Flickr images could provide a broader perspective. Together, these efforts will help refine our understanding of visual interestingness and its contributing factors.


\clearpage \clearpage
{\small
{\bf Acknowledgements.} This research was funded by the Swiss National Science Foundation (SNSF) under grant number 206319 ``Visual Interestingness -- All images are equal but some images are more equal than others''.}

{\small
\bibliographystyle{ieee_fullname}
\bibliography{egbib}
}



\end{document}